\documentclass[11pt,a4paper]{article}

\usepackage[hyperref]{emnlp-ijcnlp-2019}
\usepackage{times}
\usepackage{latexsym}
\usepackage{listings}
\usepackage{url}
\usepackage{enumitem}
\usepackage{times}
\usepackage{latexsym}
\usepackage{amssymb}
\usepackage{diagbox}
\usepackage{url}

\usepackage{subfig}
\usepackage{graphicx}
\usepackage{booktabs}
\usepackage{graphicx}
\graphicspath{ {./} }
\usepackage{multirow}

\newcommand{\nop}[1]{}

\aclfinalcopy % Uncomment this line for the final submission

\title{Unicoder: A Universal Language Encoder by Pre-training with Multiple Cross-lingual Tasks}

\author{Haoyang Huang~$^\S$ \quad Yaobo Liang~$^\S$ \quad Nan Duan$^\S$ \quad Ming Gong$^\dag$ \quad Linjun Shou$^\dag$ \\ \textbf{Daxin Jiang}$^\dag$ \quad \textbf{Ming Zhou}$^\S$ \\
  { $^\S$ Microsoft Research Asia, Beijing, China} \\
  {$^\dag$ STCA NLP Group, Microsoft, Beijing, China}\\
  {\small \tt \{haohua,yalia,nanduan,migon,lisho,djiang,mingzhou\}@microsoft.com}\\
}

\date{}

\begin{document}
\maketitle
\begin{abstract}
We present Unicoder, a universal language encoder that is insensitive to different languages. Given an arbitrary NLP task, a model can be trained with Unicoder using training data in one language and directly applied to inputs of the same task in other languages.
Comparing to similar efforts such as Multilingual BERT \cite{bert} and XLM \cite{xlm}, three new cross-lingual pre-training tasks are proposed, including cross-lingual word recovery, cross-lingual paraphrase classification and cross-lingual masked language model. These tasks help Unicoder learn the mappings among different languages from more perspectives. We also find that doing fine-tuning on multiple languages together can bring further improvement.
Experiments are performed on two tasks: cross-lingual natural language inference (XNLI) and cross-lingual question answering (XQA), where XLM is our baseline. On XNLI, 1.8\% averaged accuracy improvement (on 15 languages) is obtained. On XQA, which is a new cross-lingual dataset built by us, 5.5\% averaged accuracy improvement (on French and German) is obtained.

\end{abstract}

\section{Introduction}

Data annotation is expensive and time-consuming for most of NLP tasks. Recently, pre-trained models, such as ELMo \cite{elmo}, BERT \cite{bert} and GPT \cite{gpt}, have shown strong capabilities of transferring knowledge learned from large-scale text corpus to specific NLP tasks with limited or no training data. But they still cannot handle tasks when training and test instances are in different languages.

Motivated by this issue, some efforts have been made, such as Multilingual BERT \cite{bert} and XLM \cite{xlm}, for cross-lingual tasks. Multilingual BERT trains a BERT model based on multilingual Wikipedia, which covers 104 languages. As its vocabulary contains tokens from all languages, Multilingual BERT can be used to cross-lingual tasks directly. XLM further improves Multilingual BERT by introducing a translation language model (TLM). TLM takes a concatenation of a bilingual sentence pair as input and performs masked language model based on it. By doing this, it learns the mappings among different languages and performs good on the XNLI dataset.

However, XLM only uses a single cross-lingual task during pre-training. At the same time, \citet{liu2019improving} has shown that multi-task learning can further improve a BERT-style pre-trained model. So we think more cross-lingual tasks could further improve the resulting pre-trained model for cross-lingual tasks. To verify this, we propose Unicoder, a universal language encoder that is insensitive to different languages and pre-trained based on 5 pre-training tasks. Besides masked language model and translation language model, 3 new cross-lingual pre-training tasks are used in the pre-training procedure, including cross-lingual word recovery, cross-lingual paraphrase classification and cross-lingual masked language model. Cross-lingual word recovery leverage attention matrix between bilingual sentence pair to learn the cross-lingual word alignment relation. Cross-lingual paraphrase classification takes two sentences from different languages and classify whether they have same meaning. This task could learn the cross-lingual sentence alignment relation. Inspired by the successful of monolingual pre-training on long text~\cite{gpt,bert}, we propose cross-lingual masked language model whose input is document written by multiple languages. % Previous work ~\cite{bert,gpt} proved using document is better than sentence in language model pre-training.
We also find that doing fine-tuning on multiple languages together can bring further improvement. For the languages without training data, we use machine translated data from rich-resource languages.

Experiments are performed on cross-lingual natural language inference (XNLI) and cross-lingual question answering (XQA), where both Multilingual BERT and XLM are considered as our baselines. On XNLI, 1.8\% averaged accuracy improvement (on 15 languages) is obtained. On XQA, which is a new cross-lingual dataset built by us, 5.5\% averaged accuracy improvement (on French and German) is obtained.

In short, our contributions are 4-fold. First, 3 new cross-lingual pre-training tasks are proposed, which can help to learn a better language-independent encoder. Second, a cross-lingual question answering (XQA) dataset is built, which can be used as a new cross-lingual benchmark dataset. Third, we verify that by fine-tuning multiple languages together, significant improvements can be obtained. Fourth, on the XNLI dataset, new state-of-the-art results are achieved.

\section{Related work}
% Pretraining and multi-lingual bert, xlm
\paragraph{Monolingual Pre-training}  Recently, pretraining an encoder by language model~\cite{gpt,elmo,bert} and machine translation~\cite{mccann2017cove} have shown significant improvement on various natural language understanding (NLU) tasks, like tasks in GLUE~\cite{glue}. The application scheme is to fine-tune the pre-trained encoder on single sentence classification task or sequential labeling task. If the tasks have multiple inputs, just concatenate them into one sentence. This approach enables one model to be generalized to different language understanding tasks. Our approach also is contextual pre-training so it could been applied to various NLU tasks.

\paragraph{Cross-lingual Pre-training}  

Cross-lingual Pre-training is a kind of transfer learning with different source and target domain ~\cite{pan2010a}.
A high-quality cross-lingual
representation space is assumed to effectively perform the
cross-lingual transfer.  ~\citet{mikolov2013exploiting} has been applied small dictionaries to align
word representations from different languages and it is sufficient to align different languages with orthogonal transformation ~\cite{xing2015normalized}, even without parallel data ~\cite{lample2018word}. 
Following the line of previous work, the word alignment also could be applied to multiple languages ~\cite{ammar2016massively}. ~\citet{artetxe2018massively} use multilingual machine translation to train a multilingual sentence encoder and use this fixed sentence embedding to classify XNLI. We
take these ideas one step further by producing a pre-trained encoder instead of word embedding or sentence embedding.

Our work is based on two recent pre-trained cross-lingual encoders: multilingual BERT \footnote{https://github.com/google-research/bert/blob/master/multilingual.md} ~\cite{bert} and XLM ~\cite{xlm}. Multilingual BERT trains masked language model (MLM) with sharing vocabulary and weight for all 104 languages. But each training sample is monolingual document. Keeping the same setting, XLM proposed a new task TLM, which uses a concatenation of the parallel sentences into one sample for masked language modeling. Besides these two tasks, we proposed three new cross-lingual pre-training tasks for building a better language-independent encoder.

\section{Approach}

This section will describe details of Unicoder, including tasks used in the pre-training procedure and its fine-tuning strategy.

\subsection{Model Structure}
Unicoder follows the network structure of XLM ~\cite{xlm}. A shared vocabulary is constructed by running the Byte Pair Encoding (BPE) algorithm ~\cite{bpe_sennrich2016neural} on corpus of all languages. We also down sample the rich-resource languages corpus, to prevent words of target languages from being split too much at the character level.

\begin{figure*}[ht!]
    \centering
    \includegraphics[width=\linewidth]{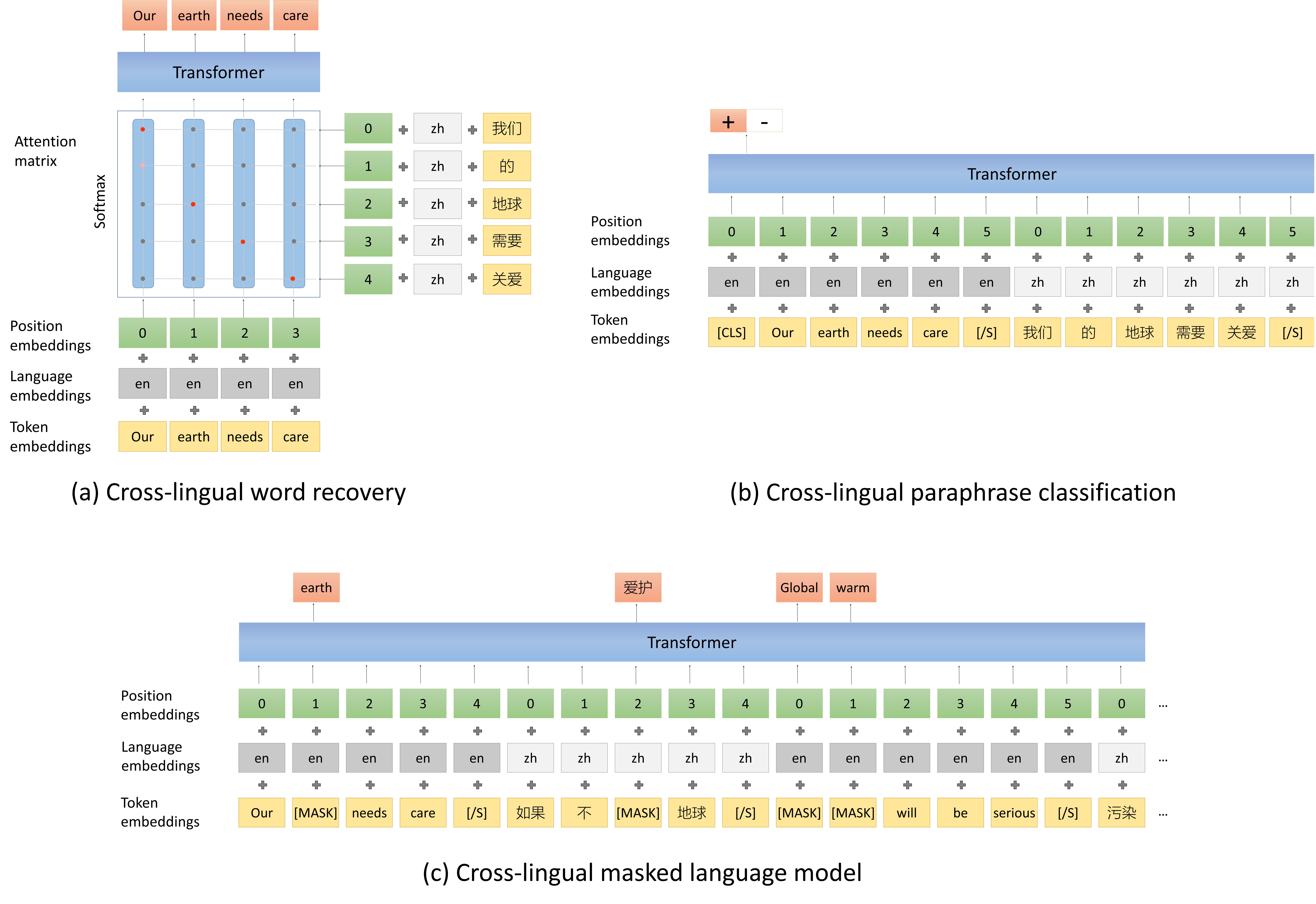}

	\caption{Unicoder consists of three cross-lingual pre-training tasks: (a) The cross-lingual word recovery model is to learn word relation from different languages (b) The cross-lingual paraphrase classification is to classify whether two sentences from different languages are paraphrase.
    (c) The cross-lingual masked language model is to train masked language model with cross-lingual document. }
	\label{fig:model_arch}
\end{figure*}

\subsection{Pre-training Tasks in Unicoder}
Both masked language model and translation language model are used in Unicoder by default, as they have shown strong performance in XLM.

Motivated by \citet{liu2019improving}, which shows that pre-trained models can be further improved by involving more tasks in pre-training, we introduce three new cross-lingual tasks in Unicoder. All training data for these three tasks are acquired from the existing large-scale high-quality machine translation corpus.

\paragraph{Cross-lingual Word Recovery} Similar to translation language model, this task also aims to let the pre-trained model learn the underlying word alignments between two languages. It is mainly motivated by the attention matrix used in the neural machine translation \citet{bahdanau2014neural} task.

Formally, given a bilingual sentence pair $(X, Y)$, where $X=(x_1,x_2,...,x_m)$ is a sentence with $m$ words from language $s$, $Y=(y_1,y_2,...,y_n)$ is a sentence with $n$ words from language $t$, this task first represents each $x_i$ as $x_i^t\in R^{h}$ by all word embeddings of $Y$:
\begin{equation}
    x^t_i=\sum_{j=1}^n{softmax(A_{ij})y^t_j} \nonumber
\end{equation}
where $x_i^s\in R^{h}$ and $y_j^t\in R^{h}$ denote the word embeddings of $x_i$ and $y_j$ respevtively, $h$ denotes the word embedding dimension, $A\in R^{m\times n}$ is an attention matrix calculated by:
\begin{equation}
A_{ij}=W[x^s_i,y^t_j,x^s_i\odot y^t_j] \nonumber
\end{equation}
$W\in R^{3*h} $ is a trainable weight and $\odot$ is element-wise multiplication. 
Then, Unicoder takes $X^t=(x^t_1,x^t_2,...,x^t_n)$ as input, and tries to predict the original word sequence $X$.

Similar to translation language model in XLM, this task is based on the bilingual sentence pairs as well. However, as it doesn't use the original words as input, we can train this task by recovering all words at the same time. The model structure of this task is illustrated in Figure 1.a.

\paragraph{Cross-lingual Paraphrase Classification}

This task takes two sentences from different languages as input and classifies whether they are with the same meaning. Like the next sentence prediction task in BERT, we concatenate two sentences as a sequence and input it to Unicoder. The representation of the first token in the final layer will be used for the paraphrase classification task. This procedure is illustrated in Figure 1.b.

We created the cross-lingual paraphrase classification dataset from machine translation dataset. Each bilingual sentence pair $(X, Y)$ servers as a positive sample. For negative samples, the most straight forward method is to replace $Y$ to a random sampled sentence from target language. But this will make the classification task too easy. So we introduce the hard negative samples followed \citet{guo2018effective}. First, we train a light-weight paraphrase model with random negative samples. Then we use this model to select sentence with high similarity score to $X$ but doesn't equal to $Y$ as hard negative samples. We choose DAN~\cite{iyyer2015deep} as the light model. We create positive and negative samples in 1:1.

\paragraph{Cross-lingual Masked Language Model}
Previous successful pre-training language model ~\cite{bert,gpt} is conducted on document-level corpus rather than sentence-level corpus. The language model perplexity on document also is much lower than sentence~\cite{elmo}. So we propose cross-lingual masked language model, whose input is come from cross-lingual document.

Cross-lingual document is a sequence of sentences, and the sentences are written with different languages. In most case, people won't write cross-lingual document. %So we generate it from machine translation data.
But we found that a large proportion of aligned sentence pairs in machine translation are extracted from parallel documents, such as MultiUN corpus and OpenSubtitles corpus.
In other words, these MT corpus are document-level corpus in which each sentence and its translation is well aligned. We construct cross-lingual document by replacing the sentences with even index to its translation as illustrated in Figure 1.c. We truncate the cross-lingual document by 256 sequence length and feed it to Unicoder for masked language modeling.

\subsection{Multi-language Fine-tuning}

A typical setting of cross-lingual language understanding is only one language has training data, but the test is conducted on other languages. We denote the language has training data as source language, and other languages as target languages. A scalable way ~\cite{conneau2018xnli} to address this problem is through Cross-lingual TEST, in which a pre-trained encoder is trained on data in source language and directly evaluated on data in target languages.

There are two other machine translation methods that make training and test belong to the same language. TRANSLATE-TRAIN translates the source language training data to a target language and fine-tunes on this pseudo training data. TRANSLATE-TEST fine-tunes on source language training data, but translates the target language test data to source language and test on it. % Previous work ~\cite{xlm} shows that TRANSLATE-TRAIN has better performance. 

Inspired by multi-task learning ~\cite{liu2018empower,liu2019improving} for improving pre-trained model, we propose a new fine-tuning strategy Multi-language Fine-tuning. We propose to fine-tune on both the source language training data and pseudo target language training data. If there are multiple target languages, we will fine-tune on all of them at same time.

% The success of multi-task learning re

Different languages may have totally different vocabulary and syntax. But our experiments show that in most cases, joint fine-tuning multiple languages could bring huge improvement. Only in just a few cases, this may harm the performance.

\begin{figure}[ht!]
    \centering
    \includegraphics[width=\linewidth]{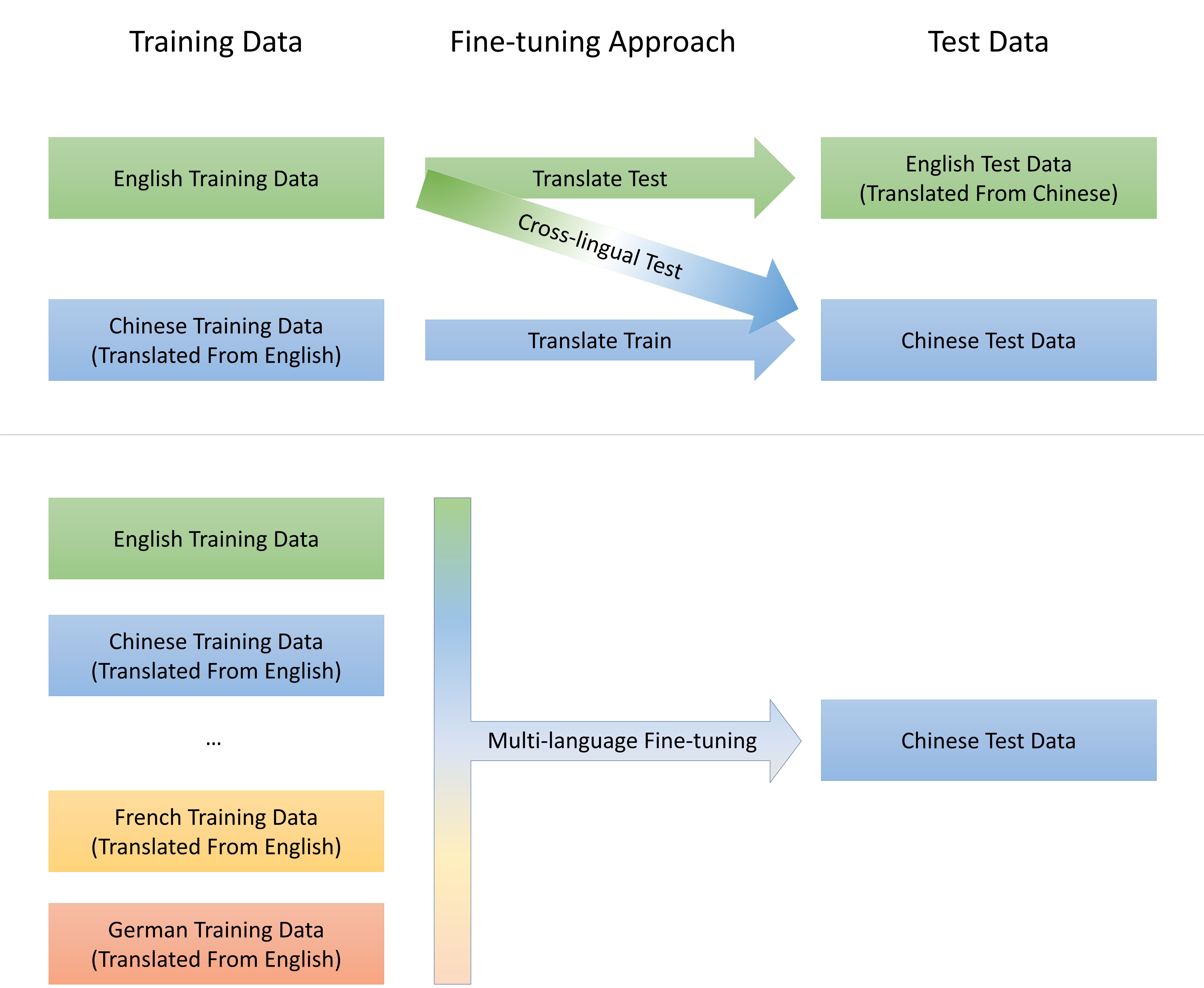}
	\caption{Currently cross-lingual fine-tuning has three baseline approaches, they could be defined based on their training data and test data. Suppose we target to test on Chinese data, Translate-train is to train on Chinese training data which is translated from English and test on Chinese test data; Translate-Test is to train on English training data and test on English test data which is translated from Chinese; Cross-lingual test is to train in English training data and test on Chinese test data. Multi-language fine-tuning is to train on English training data and multiple other languages training data which are translated from English, then test on Chinese Test data.}
	\label{fig:fine-tune}
\end{figure}

\section{Experiment}

In this section, we describe the data processing and training details. Then we compare the Unicoder with the current state of the art approaches on two tasks: XNLI and XQA.

\subsection{Data Processing}
Our model is pre-trained on 15 languages, including English(en), French(fr), Spanish(es), German(de), Greek(el), Bulgarian(bg), Russian(ru), Turkish(tr), Arabic(ar), Vietnamese(vi), Thai(th), Chinese(zh), Hindi(hi), Swahili(sw) and Urdu(ur). For MLM, we use the Wikipedia from these languages. The other four tasks need MT dataset. We use same MT dataset as ~\citet{xlm} which are collected from MultiUN \cite{multiun}, IIT Bombay corpus \cite{kunchukuttan2017the}, OpenSubtitles  2018, EUbookshop corpus and GlobalVoices. In the MT corpus, 13 of 14 languages (except IIT Bombay corpus) are from parallel document and could be used to train cross-lingual document language model. The number of data we used is reported at table ~\ref{data-number}.

For tokenization, we follows the line of \citet{koehn2007moses,chang2008optimizing} for each language. We use byte-pair encoding (BPE) to process the corpus and build vocabulary.

\begin{table}[h]
    \small
    \begin{center}
         % \resizebox{1\linewidth}{!}{
    	\begin{tabular}{l|cc}
        	\toprule
        	language & mono-lingual & bi-lingual 
        	\\
            \midrule
                ar & 3.8M & 9.8M \\
                bg & 1.5M & 0.6M \\
                de & 17.4M & 9.3M \\
                el & 1.3M & 4.0M \\
                en & 43.2M & - \\
                es & 11.3M & 11.4M \\
                fr & 15.5M & 13.2M \\
                hi & 0.6M & 1.6M \\
                ru & 12.6M & 11.7M \\
                sw & 0.2M & 0.2M \\
                th & 0.8M & 3.3M \\
                tr & 1.8M & 0.5M \\
                ur & 0.5M & 0.7M \\
                vi & 3.8M & 3.5M \\
                zh & 5.5M & 9.6M \\
        	\bottomrule
        \end{tabular}
        % }
        \end{center}
    \caption{\label{data-number} Sentence number we used in pre-training.}
\end{table}

\subsection{Training Details}
\paragraph{Model Structure} Our Unicoder is a 12-layer transformer with 1024 hidden units, 16 heads, GELU activation \cite{hendrycks2017bridging}. we set dropout to 0.1. The vocabulary size is 95,000.
\paragraph{Pre-training details}
To reduce pre-training time, we initialize our model from XLM ~\cite{xlm}. We pretrain Unicoder with five tasks including MLM$\backslash$TLM and our three cross-lingual tasks. 
% In every step, we sample a batch of data from 15 languages and 5 tasks with equal probability.
In each step, we iteratively train these five tasks. A batch for these tasks is available in 15 languages, and we sample several languages with equal probability.
And we use batch size 512 by gradient accumulation. We train our model with the Adam optimizer \cite{kingma2015adam}, and learning rate starts from $1e-5$ with invert square root decay ~\cite{transformer}. We run our pretraining experiments on a single server with 8 V100 GPUs and use FP16 to save the memory. 

The max sequence length of MLM and cross-lingual language model is 256. For the other three tasks with two sentences as input, we set the max sequence length to 128 so the sum of them is 256.

\paragraph{Fine-tuning details}
For fine-tuning stage, we use same optimizer and learning rate as pre-training. We set the batch size to 32. %We select the best result on development set and report the accuracy on test set.

\subsection{Experimental evaluation}

\begin{table*}[ht]
    \begin{center}
            \resizebox{1\linewidth}{!}{
    	\begin{tabular}{l|ccccccccccccccc|c}
        	\toprule
        	& en & fr & es & de & el & bg & ru & tr & ar & vi & th & zh & hi & sw & ur & average \\
            \midrule
            \multicolumn{17}{l}{\it Machine translate at training (TRANSLATE-TRAIN)} \\
            \midrule
            \citet{conneau2018xnli} & 73.7 & 68.3 & 68.8 & 66.5 & 66.4 & 67.4 & 66.5 & 64.5 & 65.8 & 66.0 & 62.8 & 67.0 & 62.1 & 58.2 & 56.6 & 65.4 \\
            Multilingual BERT \cite{bert} & 81.9 & - & 77.8 & 75.9 & - & - & - & - & 70.7 & - & - & 76.6 & - & - & 61.6 & - \\
            Multilingual BERT from ~\citealt{mBert} & 82.1 & 76.9 & 78.5 & 74.8 & 72.1 & 75.4 & 74.3 & 70.6 & 70.8 & 67.8 & 63.2 & 76.2 & 65.3 & 65.3 & 60.6 & 71.6 \\
            XLM \cite{xlm} & 85.0 & 80.2 & 80.8 & 80.3 & 78.1 & 79.3 & 78.1 & 74.7 & 76.5 & 76.6 & 75.5 & 78.6 & 72.3 & 70.9 & 63.2 & 76.7 \\
            
            Unicoder & 85.1 & 80.0 & 81.1 & 79.9 & 77.7 & 80.2 & 77.9 & 75.3 & 76.7 & 76.4 & 75.2 & 79.4 & 71.8 & 71.8 & 64.5 & 76.9 \\
        	\midrule
        	
        	\multicolumn{17}{l}{\it Machine translate at test (TRANSLATE-TEST)} \\
        	\midrule
        	\citet{conneau2018xnli} & 73.7 & 70.4 & 70.7 & 68.7 & 69.1 & 70.4 & 67.8 & 66.3 & 66.8 & 66.5 & 64.4 & 68.3 & 64.2 & 61.8 & 59.3 & 67.2 \\
        	Multilingual BERT \cite{bert} & 81.4 & - & 74.9 & 74.4 & - & - & - & - & 70.4 & - & - & 70.1 & - & - & 62.1 & - \\
        	
        	XLM \cite{xlm} & 85.0 & 79.0 & 79.5 & 78.1 & 77.8 & 77.6 & 75.5 & 73.7 & 73.7 & 70.8 & 70.4 & 73.6 & 69.0 & 64.7 & 65.1 & 74.2  \\
        	
        	Unicoder & 85.1 & 80.1 & 80.3 & 78.2 & 77.5 & 78.0 & 76.2 & 73.3 & 73.9 & 72.8 & 71.6 & 74.1 & 70.3 & 65.2 & 66.3 & 74.9 \\
        	
        	\midrule 
        	\multicolumn{17}{l}{\it Evaluation of cross-lingual sentence encoders (Cross-lingual TEST)} \\
        	\midrule 
        	\citet{conneau2018xnli} & 73.7 & 67.7 & 68.7 & 67.7 & 68.9 & 67.9 & 65.4 & 64.2 & 64.8 & 66.4 & 64.1 & 65.8 & 64.1 & 55.7 & 58.4 & 65.6 \\
        	Multilingual BERT \cite{bert} & 81.4 & - & 74.3 & 70.5 & - & - & - & - & 62.1 & - & - & 63.8 & - & - & 58.3 & - \\
        	Multilingual BERT from ~\citealt{mBert} & 82.1 & 73.8 & 74.3 & 71.1 & 66.4 & 68.9 & 69 & 61.6 & 64.9 & 69.5 & 55.8 & 69.3 & 60.0 & 50.4 & 58.0 & 66.3 \\
        	\citet{artetxe2018massively} & 73.9 & 71.9 & 72.9 & 72.6 & 73.1 & 74.2 & 71.5 & 69.7 & 71.4 & 72.0 & 69.2 & 71.4 & 65.5 & 62.2 & 61.0 & 70.2 \\
        	XLM \cite{xlm} & 85.0 & 78.7 & 78.9 & 77.8 & 76.6 & 77.4 & 75.3 & 72.5 & 73.1 & 76.1 & 73.2 & 76.5 & 69.6 & 68.4 & 67.3 & 75.1 \\
        	Unicoder & 85.1 & 79.0 & 79.4 & 77.8 & 77.2 & 77.2 & 76.3 & 72.8 & 73.5 & 76.4 & 73.6 & 76.2 & 69.4 & 69.7 & 66.7 & 75.4 \\
        	\midrule 
        	
        	\multicolumn{17}{l}{\it Multi-language Fine-tuning} \\
        	\midrule 
        	XLM \cite{xlm} & 85.0 & 80.8 & 81.3 & 80.3 & 79.1 & 80.9 & 78.3 & 75.6 & 77.6 & 78.5 & 76.0 & 79.5 & 72.9 & 72.8 & 68.5 & 77.8 \\
		    Unicoder w/o Word Recovery  & 85.2 & 80.5 & 81.8 & 80.9 & 79.7 & 81.1 & 79.3 & 76.2 & 78.2 & 78.5 & 76.4 & 79.7 & 73.4 & 73.6 & 68.8 & 78.2 \\
		    Unicoder w/o Paraphrase Classification  & 85.5 & 81.1 & 82.0 & 81.1 & 80.0 & 81.3 & 79.6 & 76.6 & 78.2 & 78.2 & 75.9 & 79.9 & 73.7 & 74.2 & 69.3 & 78.4 \\
		    Unicoder w/o Cross-lingual Language Model & 85.5 & 81.9 & 81.8 & 80.5 & 80.5 & 81.0 & 79.3 & 76.4 & 78.1 & 78.3 & 76.3 & 79.6 & 72.9 & 73.0 & 68.7 & 78.3 \\
        	
		    Unicoder & \textbf{85.6} & \textbf{81.1} & \textbf{82.3} & \textbf{80.9} & \textbf{79.5} & \textbf{81.4} & \textbf{79.7} & \textbf{76.8} & \textbf{78.2} & \textbf{77.9} & \textbf{77.1} & \textbf{80.5} & \textbf{73.4} & \textbf{73.8} & \textbf{69.6} & \textbf{78.5} \\

        	\bottomrule
        \end{tabular}
        }
        \end{center}
    \caption{\label{XNLI-result} Test accuracy on the 15 XNLI languages. This table is organized by fine-tuning and test approaches. TRANSLATE-TRAIN is to machine translate English training data to target language and fine-tune with this translated data; TRANSLATE-TEST is machine translate target language test data to English, the fine-tuning is conducted on English; Cross-lingual TEST is to fine-tune on English and directly test on target language; Multi-language Fine-tune is to fine-tune on machine translated training data on all languages.}
\end{table*}

\paragraph{XNLI: Cross-lingual Natural Language Inference} Natural Language Inference(NLI) takes two sentences as input and determines whether one entails the other, contradicts it or neither (neutral). XNLI is NLI defined on 15 languages. Each language contains 5000 human annotated development and test set. Only English has training data, which is a crowd-sourced collection of 433k sentence pairs from MultiNLI ~\cite{williams2018a}. The performance is evaluated by classification accuracy.

We report the results of XNLI in Table ~\ref{XNLI-result}, by comparing our Unicoder model with four baselines: \citet{conneau2018xnli} uses LSTM as sentence encoder and constraints bilingual sentence pairs have similar embedding. The other baselines are pre-training based approaches. Multilingual BERT \cite{bert} is to train masked language model on multilingual Wikipedia. And \citet{artetxe2018massively} is pre-trained with machine translation model and takes the MT encoder to produce sentence embedding. XLM \cite{xlm} explores masked language model on multilingual Wikipedia, using translation language model on MT bilingual sentence pair for pre-training in addition.

Based on the result, we could find our pre-training model Unicoder obtains the best result in every fine-tuning setting. 
In TRANSLATE-TRAIN, TRANSLATE-TEST and Cross-lingual TEST, Unicoder obtains 76.9\%, 74.9\% and 75.4\% accuracy on average, respectively. In Multi-language Fine-tuning, Unicoder outperforms XLM by 0.7\%. By
translating the English training data to target language, both TRANSLATE-TRAIN and Multi-language Fine-tuning can outperform other fine-tuning approaches on average no matter what encoder is used. Our Multi-language Fine-tuning approach is even better than TRANSLATE-TRAIN. With this approach, XLM is been improved by 1.1\% and Unicoder is been improved by 1.6\%.

By Combining Unicoder and Multi-language Fine-tuning, Unicoder achieves an new state of the art with 78.5\%. It obtains 1.8\% accuracy gain compared to previous state of the art, XLM fine-tuned with TRANSLATE-TRAIN.

\paragraph{XQA: Cross-lingual Question Answering}
We proposed a new dataset XQA. Question Answering takes a question and an answer as input, then classify whether the answer is relevant to question. Each answer is a short passage. XQA contains three languages including English, French and German. Only English have training data. The training data cover various domains, such as health, tech, sports, etc. The English training data contains millions of samples and each languages has 500 test data.

\begin{table}[h]
    \begin{center}
            \resizebox{1\linewidth}{!}{
    	\begin{tabular}{l|ccc|c}
        	\toprule
        	& en & fr & de & average  \\
            \midrule
             \multicolumn{5}{l}{\it Machine translate at training (TRANSLATE-TRAIN)} \\
            \midrule
            XLM \cite{xlm} & 80.2 & 65.1 & 63.3 & 64.2  \\
            Unicoder & 81.1 & 66.2 & 66.5 & 66.4  \\
            \midrule
            \multicolumn{5}{l}{\it Evaluation of cross-lingual sentence encoders (Cross-lingual TEST)} \\
            \midrule
            XLM \cite{xlm} & 80.2 & 62.3 & 61.7 & 62.0  \\
            Unicoder & 81.1 & 64.1 & 63.7 & 63.9  \\
            \midrule
            \multicolumn{5}{l}{\it Multi-language Fine-tuning} \\
            \midrule
            BERT \cite{bert} & 76.4 & 61.6 & 64.6 & 63.1 \\
            XLM \cite{xlm} & 80.7 & 67.1 & 68.2 & 67.7  \\
            Unicoder & \textbf{81.4} & \textbf{69.3} & \textbf{70.1} & \textbf{69.7}  \\
        	\bottomrule
        \end{tabular}
        }
        \end{center}
    \caption{\label{XQA-result}  Results on the XQA. The average column is the average of fr and de result.}
\end{table}

Keeping the same experimental setup as XNLI, we also evaluate Unicoder on XQA and
set XLM fine-tuned on our dataset as baseline with their published code. We split 5K data from training dataset as development data to do model selection. The results are shown in Table ~\ref{XQA-result}. 

\begin{table*}[ht]
    \begin{center}
         % \resizebox{1\linewidth}{!}{
    	\begin{tabular}{l|ccccccc}
        	\toprule
        	language & XNLI-en   & XNLI-ar & XNLI-es  & XNLI-fr & XNLI-ru & XNLI-zh & average
        	\\
        	number & Acc[\%]   & Acc[\%] & Acc[\%]  & Acc[\%] & Acc[\%]  & Acc[\%] & Acc[\%] \\
            \midrule
            1 & 85.1 & 76.7 & 81.1 & 80.0 & 77.9 & 79.4 & 80.0 \\
            2 & 85.2 & 77.5 & 81.5 & 80.0 &
            77.6 & 80.0 & 80.3
            \\
            6 & 85.3 & 77.9 & 81.5 & 80.4 & 78.8 & 79.9 & 80.6  \\
            15 & 85.6 & 78.2 & 82.3 & 81.1 & 79.7 & 80.5 & 81.2  \\
        	\bottomrule
        \end{tabular}
        % }
        \end{center}
    \caption{\label{language_number} Experiments of fine-tuning on different number of languages. The model is evaluated on 6 languages, and the average result is at last column. The results in the last row correspond to the results in last row of Table ~\ref{XNLI-result}.
}
\end{table*}

We could find that 1) Our model outperforms XLM at every fine-tuning setting. In Multi-language Fine-tuning, we achieved 2.0\% gain. 2) With our Unicoder, Multi-language Fine-tuning approach achieved 3.3\% gain compared to TRANSLATE-TRAIN. XLM also could been improved by 3.5\%. 3) By combining Unicoder and TRANSLATE-TRAIN, we achieve best performance 69.7\% on XQA. Compared to XLM + TRANSLATE-TRAIN baseline, we have 5.5\% gain.

\section{Analysis}
In this section,
we provide ablation analysis for different variants of our
approaches and elucidate some interesting aspects of Unicoder.
Sec. 5.1 is the ablation study of each cross-lingual pre-training task. It also shows the impact of Multi-language Fine-tuning.
Sec. 5.2 explores the impact of language numbers. 
Additionally, Sec. 5.3 analyzes the relation between English and other language by joint fine-tune on two languages. Then we further explore the relation between any language pair (Sec. 5.4).

\subsection{Ablation Study}
To examine the utility of our new cross-lingual pre-training tasks, we conducted ablation study on XNLI dataset. 
For these three cross-lingual pre-training tasks, we remove them and only pre-train on other tasks. To this end, we fine-tune the Unicoder with Multi-language Fine-tuning. The results are present at Table ~\ref{XNLI-result}.

Ablation experiments for each factor showed that removing any tasks will lead to performance drop. 
Comparing Unicoder with XLM,
We can draw several conclusions from the results in Table ~\ref{XNLI-result}. First, the cross-lingual paraphrase classification has least drop compared to others and removing the word recovery task hurts performance significantly. 
For example, in the
case of XNLI, using just the Cross-lingual Language Model and Paraphrase Classification improves test accuracy on average by 0.4\%.
And integrating with Word Recovery model allows Unicoder to learn a better representation improves the average accuracy another 0.3\%.
Second,
Multi-language fine-tuning is helpful to find the relation between languages, we will analyze it below. 
Table ~\ref{XNLI-result} and Table ~\ref{XQA-result} both show it can bring a significant boost in cross-lingual language understanding performance. With the help of Multi-language fine-tuning, Unicoder is been improved by 1.6\% of accuracy on XNLI and 3.3\% on XQA.

\subsection{The relation between language number and fine-tuning performance}

In Table ~\ref{XNLI-result}, we proved that Multi-language Fine-tuning with 15 languages is better than TRANSLATE-TRAIN who only fine-tune on 1 language. In this sub-section, we try more setting to analysis the relation between language number and fine-tuning performance.

In this experiment, only English has human labeled training data, the other languages use machine translated training data from English. The experiment is conducted on 6 languages which are the languages of MT corpus Multilingual United Nations (MultiUN).

We have four settings: \textbf{1 language} is equals to TRANSLATE-TRAIN, the pre-trained model is fine-tuned on target language. \textbf{2 languages} is to fine-tune on English and target language. For English, we report the average result when fine-tune with other 5 languages, respectively. \textbf{6 languages} is to fine-tune on 6 selected languages of this experiment. \textbf{15 languages} is to fine-tune on all 15 languages our model support, and report the results on 6 languages. This setting is equals to last row of Table ~\ref{XNLI-result}.

\begin{table*}[t]
    \begin{center}
          \resizebox{1\linewidth}{!}{
    	\begin{tabular}{l|ccccccccccccccc|c}
        	\toprule
          \diagbox{Train}{Test}	& en & fr & es & de & el & bg & ru & tr & ar & vi & th & zh & hi & sw & ur & average  \\
            \midrule
en & \textbf{85.1} & 79.0 & 79.4 & 77.8 & 77.2 & 77.2 & 76.3 & 72.8 & 73.5 & 76.4 & 73.6 & 76.2 & 69.4 & 69.7 & 66.7 & 75.4\\
fr & 77.0 & \textbf{80.0} & 79.1 & 76.9 & 77.4 & 78.6 & 76.1 & 73.1 & 73.7 & 75.3 & 73.4 & 76.2 & 70.1 & 70.3 & 66.2 & 74.9\\
es & 79.6 & 78.5 & \textbf{81.1} & 77.5 & 77.9 & 78.2 & 75.8 & 73.4 & 73.9 & 75.7 & 73.4 & 76.8 & 69.5 & 70.1 & 66.8 & 75.2\\
de & 78.2 & 77.3 & 78.3 & \textbf{79.9} & 77.1 & 77.5 & 76.7 & 73.1 & 73.8 & 75.2 & 73.4 & 76.6 & 70.9 & 71.3 & 66.7 & 75.1\\
el & 76.8 & 76.0 & 77.3 & 75.1 & 77.7 & 76.1 & 74.6 & 71.3 & 74.2 & 76.2 & 74.1 & 76.8 & 70.8 & 69.2 & 68.0 & 74.3\\
bg & 75.7 & 78.7 & 78.4 & 78.0 & 77.9 & \textbf{80.2} & 77.2 & 74.0 & 74.4 & 75.5 & 73.4 & 76.6 & 70.1 & 71.0 & 67.3 & 75.2\\
ru & 76.4 & 78.7 & 79.0 & 78.3 & \textbf{78.0} & 78.8 & \textbf{77.9} & 73.6 & 75.5 & 76.3 & 74.7 & 77.5 & 71.2 & 71.2 & \textbf{68.3} & \textbf{75.7}\\
tr & 75.4 & 75.7 & 76.3 & 75.3 & 76.2 & 76.0 & 74.8 & \textbf{75.3} & 73.0 & 73.2 & 72.1 & 75.0 & 69.7 & 70.0 & 65.6 & 73.6\\
ar & 75.5 & 76.7 & 78.4 & 77.5 & 76.7 & 77.6 & 76.1 & 73.4 & \textbf{76.7} & 75.7 & 73.4 & 76.3 & 70.1 & 70.5 & 66.2 & 74.7\\
vi & 75.1 & 77.0 & 78.6 & 77.1 & 77.2 & 78.1 & 76.2 & 73.1 & 74.0 & \textbf{76.4} & 73.6 & 76.1 & 70.5 & 71.0 & 66.9 & 74.7\\
th & 75.8 & 76.0 & 76.8 & 74.3 & 75.7 & 76.5 & 74.8 & 74.8 & 73.1 & 75.4 & \textbf{75.2} & 76.1 & 68.7 & 69.1 & 67.0 & 74.0\\
zh & 78.0 & 77.5 & 78.3 & 76.6 & 76.9 & 77.4 & 75.8 & 73.4 & 73.9 & 76.1 & 75.2 & \textbf{79.4} & 69.9 & 70.8 & 67.4 & 75.1\\
hi & 76.4 & 76.5 & 77.5 & 75.9 & 76.4 & 74.9 & 72.7 & 72.9 & 73.1 & 74.3 & 72.6 & 75.7 & \textbf{71.8} & 70.1 & 67.8 & 73.9\\
sw & 76.2 & 76.0 & 76.8 & 74.6 & 75.8 & 75.8 & 74.7 & 72.4 & 73.2 & 74.4 & 72.7 & 74.8 & 70.0 & \textbf{71.8} & 66.3 & 73.7\\
ur & 72.1 & 70.5 & 71.0 & 70.5 & 70.5 & 70.2 & 69.9 & 66.9 & 67.7 & 68.8 & 69.2 & 71.5 & 67.2 & 66.9 & 64.5 & 69.2\\
        	\bottomrule
        \end{tabular}
        }
        \end{center}
    \caption{\label{relation-result}  Accuracy on the XNLI test set of when fine-tuning Unicoder with one language and testing on other languages. The results in the diagonal correspond to the TRANSLATE-TRAIN accuracy reported in Table ~\ref{XNLI-result}.}
\end{table*}

The results are shown at Table ~\ref{language_number}. In most languages, we could find that the more languages we used in fine-tuning, the better the performance. Chinese and Russian have two numbers don't follow this trend. But 15 languages always outperform 1 language for each languages.

The most surprising result is English could be improved by Multi-language Fine-tuning even it is source language and has human-labeled training data. In next experiment, we will show that in 2 languages setting, the improvement on English is not stable and depends on the another language. But from 1 language to 6 languages and to 15 languages, English has stable improvement.

\begin{table}[h]
    % \small
    \begin{center}
         % \resizebox{1\linewidth}{!}{
    	\begin{tabular}{l|cc}
        	\toprule
        	language & XNLI-en & average  
        	\\
        	pairs & Acc[\%]  &  Acc[\%] \\
            \midrule
            ar-en & 84.8 & 76.0  \\
            bg-en & 85.1 & 76.5  \\
            de-en & 84.3 & 76.1   \\
            el-en & 84.8 & 76.2  \\
            es-en & 85.3 & 76.6  \\
            fr-en & 85.4 & 76.8  \\
            hi-en & 84.4 & 76.3 \\
            ru-en & 85.2 & 76.2   \\
            sw-en & 85.0 & 75.7  \\
            th-en & 84.1 & 75.4  \\
            tr-en & 85.0 & 75.5 \\
            ur-en & 85.0 & 75.2 \\
            vi-en & 84.4 & 75.2 \\
            zh-en & 85.1 & 76.6 \\
            en & 85.1 & 75.3 \\
        	\bottomrule
        \end{tabular}
        % }
        \end{center}
    \caption{\label{en-other-language} Result of joint fine-tuning two languages. This table reports the result on English and average accuracy of 15 languages on XNLI. The last row means Unicoder only fine-tunes on English.}
\end{table}

\subsection{The relation between English and other languages}
In this experiment, we joint fine-tune English and one language. With this experiment, we could test the relation between English and other languages since all languages have equal position in the pre-training and fine-tuning. 

We report the performance on English and average of 15 languages. First, we could find most of the average results are improved by joint fine-tuning two languages. Only Vietnamese(vi) and Urdu(ur) lead to performance drop.
Secondly, the improvement on English is not stable. French(fr) and Spanish(es) could improve English performance. But Vietnamese(vi) and Thai(th) lead to a big drop.

\subsection{The relation between different languages}

So as to better understand the relation between different languages,
we fine-tune Unicoder on one language and test on all 15 languages. The results are shown in Table ~\ref{relation-result}. The numbers in the diagonal correspond to the TRANSLATE-TRAIN result reported in Table ~\ref{XNLI-result}.

We observe that the Unicoder can transfer knowledge from one language to another language. 
We could find that fine-tune on one language often lead to best performance on this language, except Greek(el) and Urdu(ur). In fact, TRANSLATE-TRAIN of Urdu even harm the performance. Urdu also have worst generation ability to other languages. 
Russian(ru) has best generalization ability, even better than source language English.

We also could find that transfer between English(en), Spanish(es) and French(fr) is easier that other languages. The MT system between these languages also outperform other languages ~\cite{conneau2018xnli}.

\section{Conclusion}
We have introduced the Unicoder which is insensitive to different languages. We pre-train Unicoder with three new cross-lingual tasks, including cross-lingual word recovery, cross-lingual paraphrase classification and cross-lingual masked language model. We also proposed a new Multi-language Fine-tuning approach. The experiments on XNLI and XQA proved Unicoder could bring large improvements. and our approach become new state of the art on XNLI and XQA. We also did experiments to show that the more languages we used in fine-tuning, the better the results. Even rich-resource language also could been improved.

\bibliography{emnlp-ijcnlp-2019}
\bibliographystyle{acl_natbib}

\end{document}